\documentclass[letterpaper]{article} 
\usepackage{aaai2026}  
\usepackage{times}  
\usepackage{helvet}  
\usepackage{courier}  
\usepackage[hyphens]{url}  
\usepackage{graphicx} 
\usepackage{natbib}  
\usepackage{caption} 
\usepackage{algorithm}
\usepackage{booktabs} 
\usepackage{siunitx}  
\usepackage[skip=0.5\baselineskip]{caption} 
 \usepackage{amsmath}
 \usepackage{amsmath}
\usepackage{amssymb}
\usepackage{amsfonts}
\usepackage{graphicx}

\usepackage{algorithm}
\usepackage{algpseudocode}
\usepackage{newfloat}
\usepackage{listings}
\DeclareCaptionStyle{ruled}{labelfont=normalfont,labelsep=colon,strut=off} 
\floatstyle{ruled}
\newfloat{listing}{tb}{lst}{}
\floatname{listing}{Listing}
\title{HiCL: Hippocampal-Inspired Continual Learning}

\author {
    Kushal Kapoor,\textsuperscript{\rm 1}
    Wyatt Mackey,\textsuperscript{\rm 2}
    Yiannis Aloimonos,\textsuperscript{\rm 1}
    Xiaomin Lin\textsuperscript{\rm 1,3,4}
}
\affiliations {
    \textsuperscript{\rm 1}Perception and Robotics Group, University of Maryland\\
    \textsuperscript{\rm 2}DEVCOM Army Research Lab (ARL)\\
    \textsuperscript{\rm 3}Department of Electrical and Computer Engineering, Johns Hopkins University \\
    \textsuperscript{\rm 4}Department of Electrical Engineering, University of South Florida\\
    kushalk@terpmail.umd.edu, wyatt.t.mackey.civ@army.mil, 
    jyaloimo@umd.edu,
    xlin2@usf.edu
}

\begin{document}

\maketitle

\begin{abstract}

We propose HiCL, a novel hippocampal-inspired dual-memory continual learning architecture designed to mitigate catastrophic forgetting by using elements inspired by the hippocampal circuitry. Our system encodes inputs through a grid-cell-like layer, followed by sparse pattern separation using a dentate gyrus-inspired module with top-$k$ sparsity. Episodic memory traces are maintained sin a CA3-like autoassociative memory. Task-specific processing is dynamically managed via a DG-gated mixture-of-experts mechanism, wherein inputs are routed to experts based on cosine similarity between their normalized sparse DG representations and learned task-specific DG prototypes computed through online exponential moving averages. This biologically grounded yet mathematically principled gating strategy enables differentiable, scalable task-routing without relying on a separate gating network, and enhances the model's adaptability and efficiency in learning multiple sequential tasks. Cortical outputs are consolidated using Elastic Weight Consolidation weighted by inter-task similarity. Crucially, we incorporate prioritized replay of stored patterns to reinforce essential past experiences. Evaluations on standard continual learning benchmarks demonstrate the effectiveness of our architecture in reducing task interference, achieving near state-of-the-art results in continual learning tasks at lower computational costs.
\end{abstract}
\begin{links}
\link{Code}{https://github.com/kushalk173-sc/HiCL}
\link{Extended version}{https://arxiv.org/pdf/2508.16651}
\end{links}
\vspace{-5mm}
\section{Introduction}
\label{sec:intro}

Artificial neural networks excel at many single‐task benchmarks but struggle when trained sequentially on multiple tasks, suffering from \emph{catastrophic forgetting}—the tendency to overwrite previous knowledge with new information. In contrast, the human brain learns continuously throughout life, preserving decades of memories while acquiring new skills.

In continual learning, every new task is like a distinct episode in our life’s unfolding story. Just as the hippocampus rapidly encodes unique experiences into \emph{episodic memory}, tagging them with contextual and temporal cues, a continual learner must capture each task’s defining features before they fade. The brain then \emph{replays} these episodes—during sleep or quiet wakefulness—to gradually integrate them into neocortical networks, preventing older memories from being overwritten by new ones. We leverage this \emph{complementary learning systems} principle by treating each task as an episodic trace: we store compact representations in a replay buffer and periodically revisit them during training, mirroring how hippocampal replay consolidates previous and precious episodes into lasting knowledge \cite{mcclelland1995there}.
\begin{figure}[t]
    \centering
    \includegraphics[width=0.45\textwidth]{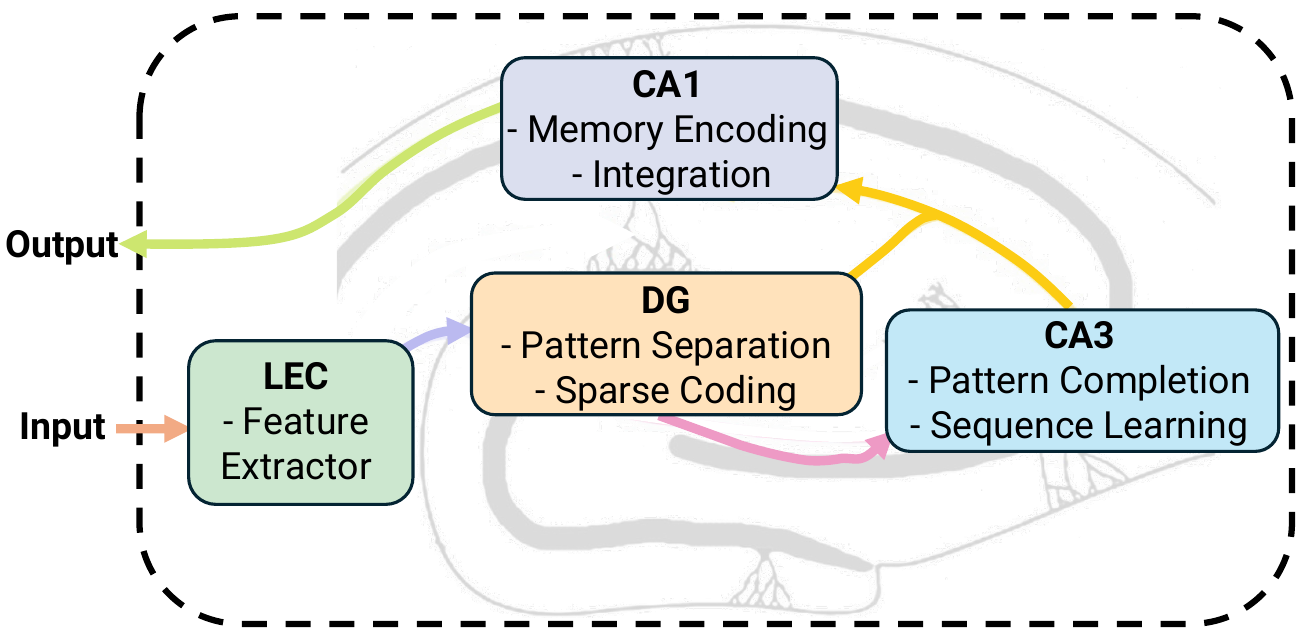}
    \caption{Schematic of the \textbf{HiCl architecture} overlaid with the hippocampal trisynaptic circuit. The entorhinal cortex (layers II/III) projects to the dentate gyrus (DG), which performs pattern separation. DG outputs feed into CA3, an autoassociative network responsible for pattern completion. CA3 then projects to CA1, which integrates inputs from both CA3 and the entorhinal cortex before sending output to other brain regions. This circuit underlies memory encoding and retrieval processes. The general background here of the hippocampal circuit is based on \cite{petrantonakis2014compressed} and adapted from \citet{moser2008place, leutgeb2007pattern, rolls2013pattern, lisman2005recall}.
}\vspace{-5mm}
    \label{fig:hippocampal_flow_intro}
\end{figure}

We draw on three core hippocampal subregions—the dentate gyrus (DG), CA3, and CA1—to inspire a \emph{DG‑Gated Mixture‑of‑Experts (MoE)} architecture(as shown in Figure~\ref{fig:hippocampal_flow_intro}, where each subcircuit is instantiated as an expert and inputs are dynamically routed based on their DG activations. This DG‑based gating, eliminates the need for task labels at inference and directly tackles catastrophic forgetting.

\textbf{Dentate Gyrus (DG):} In biology, the DG performs \emph{pattern separation} by mapping inputs onto a very sparse, typically only $3–5\%$ of granule cells fire for any given pattern \cite{hainmueller2020dentate}. In our model, the \texttt{SparseActivation} DG layer enforces top‑$k$ sparsity $(k = 5\%)$, orthogonalizing features both for expert representations and for routing signals.

\textbf{CA3:} The DG’s orthogonal codes feed into CA3’s recurrent attractor network, which functions as \emph{pattern completion}, reconstructing full memory traces from partial cues \cite{gilbert2009role}. We mirror this with a lightweight two-layer MLP that refines and transforms DG outputs via a non-linear projection, functionally completing the pattern for downstream processing.

\textbf{CA1:} Finally, CA1 integrates completed patterns from CA3 with direct cortical inputs, acting as a comparator and coordinating memory consolidation through hippocampo‑cortical dialogue \cite{bartsch2011ca1}. We emulate this via our \texttt{ca1\_integration} block and a consolidation stage combining Elastic Weight Consolidation (EWC) with a replay buffer.

\textbf{Bringing it together:} By instantiating \(N\) such experts and routing via cosine similarities between live DG codes and stored DG prototypes---using either hard or soft‑max gating---and then sharpening DG separation with a contrastive tuning phase.

This biology‑inspired solution to the stability–plasticity dilemma offers actionable insights for continual learning in artificial systems. Our \emph{DG‑Gated Mixture‑of‑Experts} architecture translates hippocampal principles into three core mechanisms: (1) a top-$k$ sparse DG layer ($k\approx5\%$) that orthogonalizes inputs for both expert encoding and dynamic routing; (2) a CA3‑inspired two‑layer MLP autoencoder that completes partial patterns via attractor-like reconstruction; and (3) a consolidation module combining a prioritized replay buffer with Elastic Weight Consolidation to protect prior knowledge. By instantiating the hippocampal tripartite circuit across multiple experts and leveraging DG‑based gating to minimize interference, CA3‑style pattern completion for robust recall, and replay+EWC for selective consolidation, our DG‑Gated MoE exemplifies how neuroscience‑informed architectures can achieve efficient, adaptive continual learning.

\begin{figure*}
    \centering
    \includegraphics[width=0.95\linewidth]{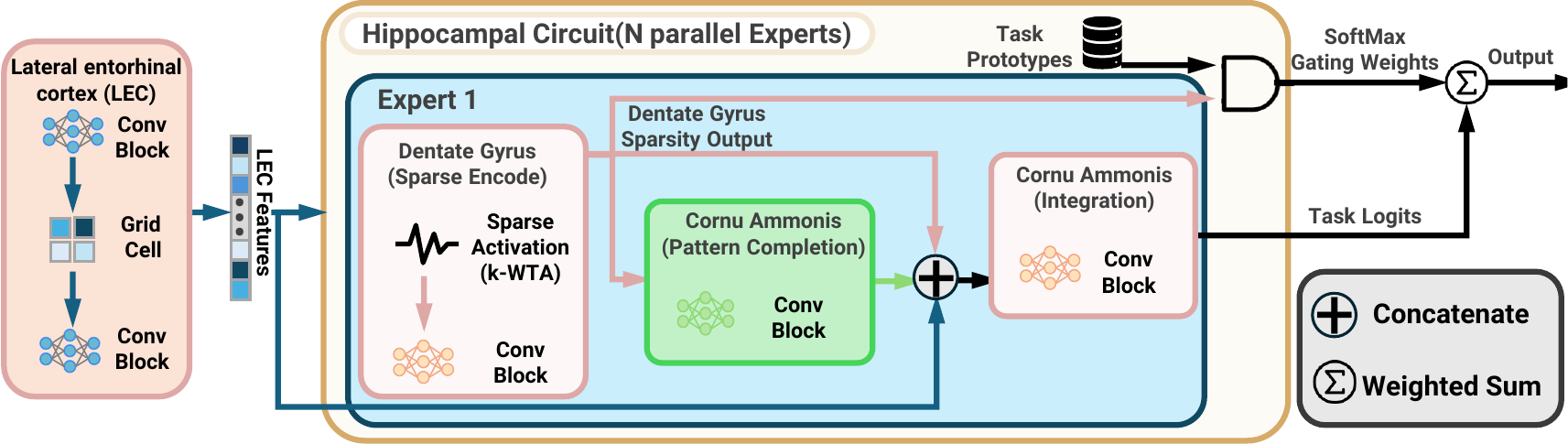}
    \caption{Overview of the HiCL architecture aligned with the hippocampal trisynaptic circuit: EC II/III → DG (pattern separation) → CA3 (pattern completion) → CA1 (integration). }
    \vspace{-5mm}
 \label{fig:hippocampal_flow_conceptual}
\end{figure*}
Our primary contributions are:
\begin{itemize}
    \item We \textbf{introduce HiCL, a DG-gated Mixture-of-Experts (MoE) model that routes inputs without a gating network.} By leveraging the sparse representations from a Dentate Gyrus-inspired layer, our model dynamically assigns tasks to experts based on cosine similarity to learned prototypes.
    \item We \textbf{developed a training strategy that explicitly improves pattern separation.} Our two-phase approach first trains individual experts using EWC, replay, and feature distillation, then applies a global contrastive loss to all DG layers, forcing them to learn more distinct representations for their respective tasks.
    \item We \textbf{achieved competitive accuracy at a fraction of the computational cost.} On benchmarks like Split CIFAR-10, our model demonstrates that a neuroscience-aligned architecture can be both computationally effective and efficient in a continual learning setting.
\end{itemize}

\section{Related Work}

Continual learning (CL) aims to develop systems that learn from a non-stationary data stream without catastrophic forgetting. In this section, we review key literature spanning computational neuroscience and machine learning, focusing on complementary learning systems, hippocampal memory motifs, and modern CL strategies.

\paragraph{Complementary Learning Systems (CLS).} The CLS theory posits a dual-memory system: a fast hippocampal module for episodic learning and a slow neocortical system for generalization~\cite{mcclelland1995there, kemker2018measuring}. This foundational concept has influenced many biologically inspired CL frameworks. Examples include hybrid cortico-hippocampal networks that combine feedforward and spiking layers~\cite{lee2020cortical}, DualNet's distinct fast/slow learners~\cite{Pham2021DualNet}, FearNet's use of generative replay~\cite{Kemker2018FearNet}, and Triple Memory Networks that model hippocampus-PFC interactions with replay and consolidation~\cite{wang2021triple}.

\paragraph{Hippocampal Replay.} Replay of past experiences is a powerful anti-forgetting mechanism, mirroring the brain's hippocampal replay during sleep and wakefulness~\cite{cairney2018memory}. In machine learning, this is realized through \textbf{experience replay}, which explicitly rehearses stored examples~\cite{rolnick2019experience}, or \textbf{generative replay}, which trains a network to synthesize pseudo-data~\cite{shin2017continual}. We adopt prioritized experience replay, inspired by findings that emphasizing important memories enhances learning efficiency in both reinforcement learning~\cite{schaul2016prioritized} and biological systems~\cite{Mattar2018prioritized}. The ability to generate contrastive data during replay is considered crucial for preventing representational collapse~\cite{hinton2022forward}.

\paragraph{Modern CL Architectures and Strategies.} Many CL frameworks are categorized as \textbf{parameter isolation}, \textbf{rehearsal based}, or \textbf{regularization based}~\cite{de2021continual}. \textbf{Regularization} methods add a loss penalty to prevent changes to weights important for past tasks, as seen in Elastic Weight Consolidation (EWC)~\cite{kirkpatrick2017overcoming}, Learning without Forgetting (LwF)~\cite{li2017learning}, and Synaptic Intelligence (SI)~\cite{Zenke2017SI}. \textbf{Parameter isolation} methods assign distinct weights to different tasks, using techniques such as network expansion in Progressive Nets~\cite{rusu2016progressive} or hard attention masks~\cite{serra2018hat, madotto2020continual, wang2024rehearsal}. \textbf{Rehearsal} methods use a buffer of past data~\cite{rebuffi2017icarl, riemer2018learning, zhang2022continual} or a generative model~\cite{shin2017continual, su2019generative, skiers2024joint}. Hybrid models such as Progress \& Compress~\cite{schwarz2018progress} and CLEAR~\cite{riemer2018learning} combine these strategies effectively, and our approach synergizes these ideas by using architectural modularity, sparsity, and replay.

\paragraph{Mixture-of-Experts (MoE) and Gating.} MoE models enable modular specialization by routing inputs to specialized subnetworks~\cite{yuksel2012twenty, shazeer2017moe}, a principle recently applied to CL~\cite{Li2025MoETheory, Le2024PromptMoE} and large-scale models~\cite{guo2025deepseek, cai2024survey}. Stability is often improved via adaptive routing, for instance by annealing a routing temperature~\cite{Nie2022EvoMoE,Fedus2022Switch}. Our DG-Gated MoE is distinct in that expert selection emerges via cosine similarity to prototypes, bypassing an explicit learned gating network.

\paragraph{Prototype-Based Replay.} Storing class prototypes instead of raw data is a memory-efficient rehearsal strategy pioneered by iCaRL~\cite{rebuffi2017icarl}. Subsequent work has combined prototype anchoring with contrastive objectives to preserve the geometry of the feature space across tasks~\cite{Caccia2022ContrastiveCL, Aghasanli2025Prototype}. We follow this line, maintaining online exponential moving average (EMA) prototypes and applying a push-pull contrastive loss during replay.

These prior threads motivate a more tightly coupled design: rather than treating biologically inspired modules as loose metaphors, we embed them directly into representation, routing, and training dynamics in a unified hippocampal-inspired dual-memory MoE.

\section{Biologically Inspired Motifs and Dual-Memory Training}
\label{sec:bio_modules}

Our design is grounded in the hippocampal trisynaptic circuit (DG \(\rightarrow\) CA3 \(\rightarrow\) CA1) and the theory of complementary learning systems (CLS)~\cite{mcclelland1995there, kemker2018measuring}. We operationalize these motifs as inductive scaffolds: structured encoding, sparse separation, nonlinear refinement/completion, and integration, which together feed into prototype-based routing and consolidation. Replay-inspired mechanisms and prototype anchoring provide retention and interference mitigation, while modular gating arises naturally from the quasi-orthogonal codes produced by DG-like separation. Learning is organized as a two-phase dual-memory schedule—specialization followed by consolidation—whose detailed losses are defined in the Architecture section.

\noindent\textbf{Dual-Memory Realization.} We operationalize the complementary learning system as a two-phase training schedule: \textbf{Phase I} rapidly specializes, forming and updating task prototypes and shaping DG gating, while \textbf{Phase II} consolidates prior knowledge via contrastive alignment, mitigating interference and preventing degradation.

\subsection{Grid Cell Encoding}
Grid-cell-like representations from the entorhinal cortex provide a structured relational prior that has been shown to enhance spatial and contextual reasoning in downstream tasks~\cite{hafting2005microstructure, Banino2018Grid}. We implement this by applying \(M=4\) parallel \(1\times1\) convolutions with learned phase offsets to backbone features \(\mathbf{f}\):
\begin{align*}
  \mathbf{g}_m &= \sin(\mathbf{W}_m\,\mathbf{f} + \phi_m),\quad m=1,\dots,4,
  \mathbf{g} = [\mathbf{g}_1; \dots; \mathbf{g}_4].
\end{align*}

This hexagonal-style embedding enhances discrimination and imposes structured relational priors before gating, similar in spirit to positional priors used in other continual and representation learning works.

\subsection{Dentate Gyrus: Sparse Pattern Separation}
To reduce interference between tasks, we draw on the role of the dentate gyrus in orthogonalizing inputs via sparse coding~\cite{leutgeb2007pattern, rolls2013cortical}. The encoded representation \(\mathbf{g}\) is projected and sparsified:
\begin{equation}
\begin{aligned}
  \mathbf{z} &= \mathrm{ReLU}(\mathbf{W}_{\mathrm{DG}}\mathbf{g} + \mathbf{b}_{\mathrm{DG}}), \\
  \mathbf{p}_{\mathrm{sep}}
    &= \mathrm{TopK}(\mathrm{LayerNorm}(\mathbf{z}),\,k),\quad
    k = \left\lfloor 0.05\,\dim(\mathbf{z})\right\rfloor.
\end{aligned}
\label{eq:dg_sparse}
\end{equation}

This top-\(k\) mechanism approximates inhibitory microcircuitry to produce quasi-orthogonal sparse codes, creating natural modularization and enabling routing without explicit task labels, akin to capacity partitioning in MoE and gating literature~\cite{yuksel2012twenty, shazeer2017moe}.

\subsection{CA3: Nonlinear Refinement / Completion}
The sparse separation codes are refined through a feed-forward MLP to approximate the pattern completion function attributed to CA3~\cite{treves1994autoassociation}. Unlike classical recurrent attractor implementations~\cite{Hopfield1982Attractor}, we use:
\begin{align*}
\mathbf{p}_{\mathrm{comp}} &= \mathrm{LN}\bigl(\mathrm{ReLU}(\mathbf{W}_2\,\mathrm{ReLU}(\mathbf{W}_1\,\mathbf{p}_{\mathrm{sep}} \\
&\quad +\mathbf{b}_1)+\mathbf{b}_2)\bigr)
\end{align*}

providing nonlinear task-specific refinement while maintaining stability in the face of non-i.i.d. data~\cite{demircigil2022sparse, krotov2016dense}.

\subsection{CA1: Integration}
Separation and completion are fused to form a consolidated representation:
\[
  \mathbf{u} = [\mathbf{p}_{\mathrm{sep}};\mathbf{p}_{\mathrm{comp}}],
\]
which serves as the basis for both prediction and prototype anchoring. This mirrors CA1’s role in integrating pattern-separated and completed signals before driving downstream systems.

\subsection{Biological Abstraction.}
Our design philosophy emphasizes \textbf{principled abstraction}, extracting core computational principles from hippocampal circuits while trading biological fidelity for computational efficiency and trainability. The biological DG employs complex inhibitory microcircuitry to enforce sparse coding~\cite{hainmueller2020dentate}. We replace this with a non-parametric TopK operation that, when combined with Phase II contrastive tuning, achieves effective pattern separation without simulating neuronal dynamics. Similarly, CA3's recurrent attractor network~\cite{treves1994autoassociation} is approximated by a feedforward MLP that captures pattern completion functionality in a single stable forward pass, avoiding the computational cost and training instability of recurrent architectures~\cite{Hopfield1982Attractor}. Our learnable GridCellLayer adapts its spatial encoding to dataset geometry rather than using fixed Fourier bases, providing representational flexibility for the continual learning setting. These design choices demonstrate that capturing biological \emph{function} enables effective continual learning without requiring literal anatomical replication.

\subsection{Prototype Maintenance and Replay Motivation}
Each expert maintains a prototype \(\mathbf{u}_i\) updated during specialization via exponential moving average of its DG separation code:
\[
  \mathbf{u}_i \leftarrow (1-\mu)\,\mathbf{u}_i + \mu\,\mathbf{p}_{\mathrm{sep}}^{(i)},\quad \mu=0.01.
\]
This echoes prototype-based rehearsal approaches~\cite{rebuffi2017icarl, Caccia2022ContrastiveCL} and supplies stable references for both gating and later consolidation. The use of replay-like alignment and contrastive objectives (detailed in Section~\ref{sec:architecture}) is inspired by hippocampal replay phenomena that reinforce important memories during offline phases~\cite{cairney2018memory, schaul2016prioritized, Mattar2018prioritized}.

\section{The Architecture}
\label{sec:architecture}

Our system is a DG-Gated Mixture-of-Experts (MoE) designed to mitigate catastrophic forgetting by partitioning capacity into \(N\) specialized experts. Each expert implements a differentiable analogue of the hippocampal trisynaptic loop (DG \(\rightarrow\) CA3 \(\rightarrow\) CA1), combining sparse separation, nonlinear refinement, and integration. Expert selection is driven by similarity between current DG codes and per-expert prototypes, enabling dynamic specialization without an explicit learned gating network.

A lightweight CNN backbone (e.g., a LeNet) maps each input to feature maps \(\mathbf{f}\). The grid-cell encoding, sparse DG separation, refinement, and integration are as described in Section~\ref{sec:bio_modules}.

\subsection{Gating \& Routing}
Each expert maintains its own DG module, enabling task-specific sparse representations. At inference, inputs are processed through all $N$ DG modules in parallel to compute routing scores via cosine similarity to prototypes. Once the expert is selected, only that expert's CA3 and CA1 modules are activated for prediction. The gating computation requires $N$ DG forward passes, included in our reported FLOPs.

Each expert’s prototype \(\mathbf{u}_i\) is updated during Phase I via EMA of its DG separation code:
\[
  \mathbf{u}_i \leftarrow (1-\mu)\,\mathbf{u}_i + \mu\,\mathbf{p}_{\mathrm{sep}}^{(i)},\quad \mu=0.01.
\]
At inference, similarity scores
\[
  s_i = \cos(\mathbf{p}_{\mathrm{sep}}^{(i)}, \mathbf{u}_i)
\]
are converted to gating weights via:
\begin{itemize}
  \item \emph{Soft gating:} \(\alpha_i \propto \exp(s_i/\tau)\).  
  \item \emph{Hard gating:} \(\alpha_i = \mathbf{1}\{i=\arg\max_j s_j\}\).  
  \item \emph{Top-2:} select top two and renormalize.  
  \item \emph{Hybrid:} softmax over top-\(k\) similarities.
\end{itemize}
Outputs from experts are aggregated according to \(\{\alpha_i\}\) to form the final prediction.

\subsection{Training Objectives}
Training proceeds in two phases.
\paragraph{Phase I: Specialized Encoding}
The classification loss is augmented with auxiliary terms. The prototype proximity loss is denoted \(\mathcal{L}_{\mathrm{intra}}\) to reflect push–pull dynamics:
\begin{align}
\mathcal{L}_{\mathrm{intra}} 
  &= \sum_{i,j}\mathbf{1}_{[y_i=y_j]}\|\mathbf{p}_i-\mathbf{p}_j\|^2 \nonumber\\
  &\quad - \lambda \sum_{i,j}\mathbf{1}_{[y_i\neq y_j]}
    \bigl[\max\{0,\,m - \|\mathbf{p}_i-\mathbf{p}_j\|\}\bigr]^2.
\end{align}
Additional losses are:
\begin{itemize}
  \item \(\mathcal{L}_{\mathrm{replay}}\): cross-entropy on replayed examples.  
  \item \(\mathcal{L}_{\mathrm{distill}}\): feature distillation aligning to frozen snapshots.  
  \item \(\mathcal{L}_{\mathrm{EWC}}\): Elastic Weight Consolidation regularizer.  
  \item \(\mathcal{L}_{\mathrm{sparsity}}\): encourages target DG activation sparsity.  
\end{itemize}
The Phase I objective combines these terms:
\begin{align}
\mathcal{L}_{\mathrm{phase1}}
  &= \mathcal{L}_{\mathrm{cls}}
     + \alpha_{\mathrm{intra}}\,\mathcal{L}_{\mathrm{intra}}
     + \alpha_{\mathrm{dist}}\,\mathcal{L}_{\mathrm{distill}}
     + \alpha_s\,\mathcal{L}_{\mathrm{sparsity}}.
\end{align}

\paragraph{Phase II: Consolidation \& Contrastive Prototypes}
After freezing non-DG parameters, a prototype-based contrastive loss reinforces current-task alignment while suppressing interference:
\begin{align}
  \mathcal{L}_{\mathrm{contrastive}}
  &= \bigl(1 - \cos(\mathbf{p}_{\mathrm{sep}}^{(t)}, \mathbf{u}_t)\bigr) \nonumber \\
  &\quad + \sum_{j \neq t}
    \bigl[\max\{0,\;\cos(\mathbf{p}_{\mathrm{sep}}^{(j)}, \mathbf{u}_j) - m\}\bigr].
\end{align}
\[
\mathcal{L}_{\mathrm{phase2}}
= \alpha_{\mathrm{contrastive}}\,\mathcal{L}_{\mathrm{contrastive}}.
\]

\paragraph{Full Objective}
The overall loss combines both phases and retention mechanisms:
\[
\mathcal{L}
= \lambda_1 \mathcal{L}_{\mathrm{phase1}}
+ \lambda_2 \mathcal{L}_{\mathrm{phase2}}
+ \lambda_3 \mathcal{L}_{\mathrm{EWC}}
+ \lambda_4 \mathcal{L}_{\mathrm{replay}}.
\]

\section{Results}
\label{sec:results}

\begin{table*}[t]
\centering
\begin{tabular}{l c c c c c c c}
\toprule
\textbf{Method} & \textbf{Buffer} &
\multicolumn{3}{c}{\textbf{Split CIFAR 10}} &
\multicolumn{3}{c}{\textbf{Split TinyImageNet}} \\
\cmidrule(lr){3-5} \cmidrule(lr){6-8}
& \textbf{Size} &
\textbf{Task IL \textuparrow} & \textbf{Class IL \textuparrow} & \textbf{FLOPs \textdownarrow} &
\textbf{Task IL \textuparrow} & \textbf{Class IL \textuparrow} & \textbf{FLOPs\textdownarrow} \\
\addlinespace
\midrule
\multicolumn{8}{l}{\textit{Upper Bound}} \\
Joint & -- & $97.18 \pm 0.12$ & $87.00 \pm 1.03$ & 219.60 & $82.04 \pm 0.3$ & $60.21 \pm 0.3$ & 850.72 \\
\midrule
\multicolumn{8}{l}{\textit{Regularization-Based Methods}} \\
SGD (Fine-tuning) & -- & $66.08 \pm 7.30$ & $19.53 \pm 0.07$ & 235.66 & $4.79 \pm 0.43$ & $0.32 \pm 0.02$ & 866.77 \\
LwF & --  & $63.89 \pm 1.54$ & $19.40 \pm 0.15$ & 235.66 & $21.43 \pm 0.28$ & $8.54 \pm 0.20$ & 866.77 \\
SI & --  & $62.16 \pm 4.91$ & $19.29 \pm 0.18$ & 219.60 & $21.40 \pm 0.19$ & $8.56 \pm 0.08$ & 850.72 \\
\addlinespace
\midrule
\multicolumn{8}{l}{\textit{Replay-Based Methods}} \\
A-GEM & 200 & $85.35 \pm 2.54$ & $19.66 \pm 0.19$ & 235.66 & $22.31 \pm 0.94$ & $8.89 \pm 0.10$ & 866.77 \\
      & 500  & $85.89 \pm 2.12$ & $19.83 \pm 0.46$ & 235.66 & $23.17 \pm 0.49$ & $8.70 \pm 0.08$ & 866.77 \\
ER & 200 & $90.07 \pm 0.83$ & $47.77 \pm 1.69$ & 235.66 & $31.20 \pm 0.10$ & $8.44 \pm 0.25$ & 866.77 \\
   & 500  & $92.62 \pm 0.28$ & $61.48 \pm 3.54$ & 235.66 & $31.74 \pm 0.29$ & $8.62 \pm 0.15$ & 866.77 \\
DER++ & 200 & $89.25 \pm 0.52$ & $56.73 \pm 1.46$ & 235.66 & $33.10 \pm 0.94$ & $8.82 \pm 0.20$ & 866.77 \\
      & 500 & $91.21 \pm 0.25$ & $66.49 \pm 1.53$ & 235.66 & $32.74 \pm 0.92$ & $8.78 \pm 0.07$ & 866.77 \\
\addlinespace
\midrule
\multicolumn{8}{l}{\textit{Architectural \& Hybrid Methods}} \\
iCaRL & 200 & $94.03 \pm 0.46$ & $47.54 \pm 3.28$ & 219.60 & $24.00 \pm 0.37$ & $8.87 \pm 0.12$ & 850.72 \\
      & 500  & $94.36 \pm 0.26$ & $48.34 \pm 1.49$ & 219.60 & $23.25 \pm 0.91$ & $8.71 \pm 0.05$ & 850.72 \\
FDR & 200  & $91.23 \pm 0.50$ & $29.51 \pm 1.95$ & 236.36 & $35.90 \pm 0.33$ & $8.38 \pm 0.10$ & 867.48 \\
    & 500 & $92.34 \pm 1.26$ & $28.73 \pm 2.82$ & 236.36 & $41.37 \pm 0.50$ & $8.56 \pm 0.12$ & 867.48 \\
HAL & 200 & $77.97 \pm 2.42$ & $34.11 \pm 5.67$ & 219.60 & $21.92 \pm 0.57$ & $8.52 \pm 0.18$ & 850.72 \\
    & 500 & $84.55 \pm 1.32$ & $41.98 \pm 3.01$ & 219.60 & $21.49 \pm 0.62$ & $8.45 \pm 0.20$ & 850.72 \\
E2Net & 200 & $93.11 \pm 0.46$ & $67.23 \pm 1.57$ & 219.60 & $40.94 \pm 0.43$ & $15.35 \pm 0.08$ & 850.72 \\
      & 500 & $93.48 \pm 0.37$ & $72.19 \pm 0.82$ & 219.60 & $45.51 \pm 0.10$ & $22.47 \pm 0.15$ & 850.72 \\

SparCL            & 200   & $85.63 \pm 0.62$ & $45.21 \pm 1.10 $ & 37.22 & $39.14 \pm 0.85$ & $8.98 \pm 0.38$ & 149.40 \\
(Sparsity: 0.25)   & 500   & $87.38 \pm 0.46$ & $50.45 \pm 1.22$ & 37.22 & $50.83 \pm 0.69$ & $10.48  \pm 0.29 $ & 149.40 \\
\addlinespace
\midrule
\multicolumn{8}{l}{\textit{Our Method}} \\
HiCL Small        & 200   & $90.24 \pm 0.84$ & $57.65 \pm 0.49$  & 38.21 & $59.16 \pm 0.37$ & $23.28 \pm 0.46$ & 152.9  \\
(Sparsity: 0.15)                    & 500  & $90.36 \pm 0.54$ & $60.68 \pm 1.05$ & 38.21 & $59.63 \pm 0.85$ & $23.51 \pm 0.91$ & 152.9  \\
HiCL Large        & 200  & $91.74 \pm 0.81$ & $60.48 \pm 0.15$ & 84.18 & $\textbf{62.76} \pm 0.26$ & $25.45 \pm 0.14$  & 336.71 \\
(Sparsity: 0.15)                  & 500  & $92.38 \pm 0.49$ & $65.58 \pm 0.45$  & 84.18 & $62.60 \pm 0.79$ & $\textbf{25.71} \pm 0.64$ & 336.71 \\
\bottomrule
\end{tabular}
\small
\caption{Comparison of accuracy (\%), and FLOPs (M) on Split CIFAR-10 and Split TinyImageNet under different memory and compute settings. The baselines have been updated based on new results for Split CIFAR-10. FLOPs are reported in millions.}
\label{tab:mega-results-flops-simple}
\end{table*}

\subsection{Benchmarks and Tasks}
We evaluate our model, HiCL, on two standard image-classification benchmarks, each split into a sequence of distinct tasks to simulate a continual learning environment.

\begin{itemize}
    \item \textbf{Split CIFAR-10}~\cite{krizhevsky2009learning}: The 10 classes are divided into 5 sequential tasks, with 2 new classes introduced in each task.
    \item \textbf{Split Tiny-ImageNet}~\cite{deng2009imagenet}: A more complex benchmark where 200 classes are split into 10 tasks of 20 classes each.
\end{itemize}
For all experiments, we follow the \textbf{task-incremental learning} protocol, where the model learns tasks sequentially. We use a ResNet backbone and a fixed-size memory buffer for all methods, where applicable to ensure fair and controlled comparisons. 

\subsection{Evaluation Metrics}
Our goal is to develop a model that learns new information effectively while forgetting as little as possible, all with high computational efficiency. To measure this, we report the following metrics:

\begin{itemize}
    \item \textbf{Accuracy (Goal: Higher is Better):} We measure final average accuracy in two scenarios.
        \begin{itemize}
            \item \textbf{Task-IL Accuracy ($A_T$):} Measures performance when the task identity is provided at test time. This evaluates the model's ability to preserve task-specific knowledge.
            \item \textbf{Class-IL Accuracy ($A_C$):} Measures performance when the task identity is unknown, forcing the model to classify among all previously seen classes. This is a more challenging and realistic measure of a model's ability to avoid catastrophic forgetting.
        \end{itemize}
    \item \textbf{Efficiency (Goal: Lower is Better):} We report the computational cost in Mega-FLOPs (MFLOPs) for a single forward pass. This metric serves as a proxy for the model's inference speed and complexity.
\end{itemize}

\begin{figure}[t]
    \centering
    \begin{minipage}{1\linewidth}
        \centering
        \includegraphics[width=\linewidth]{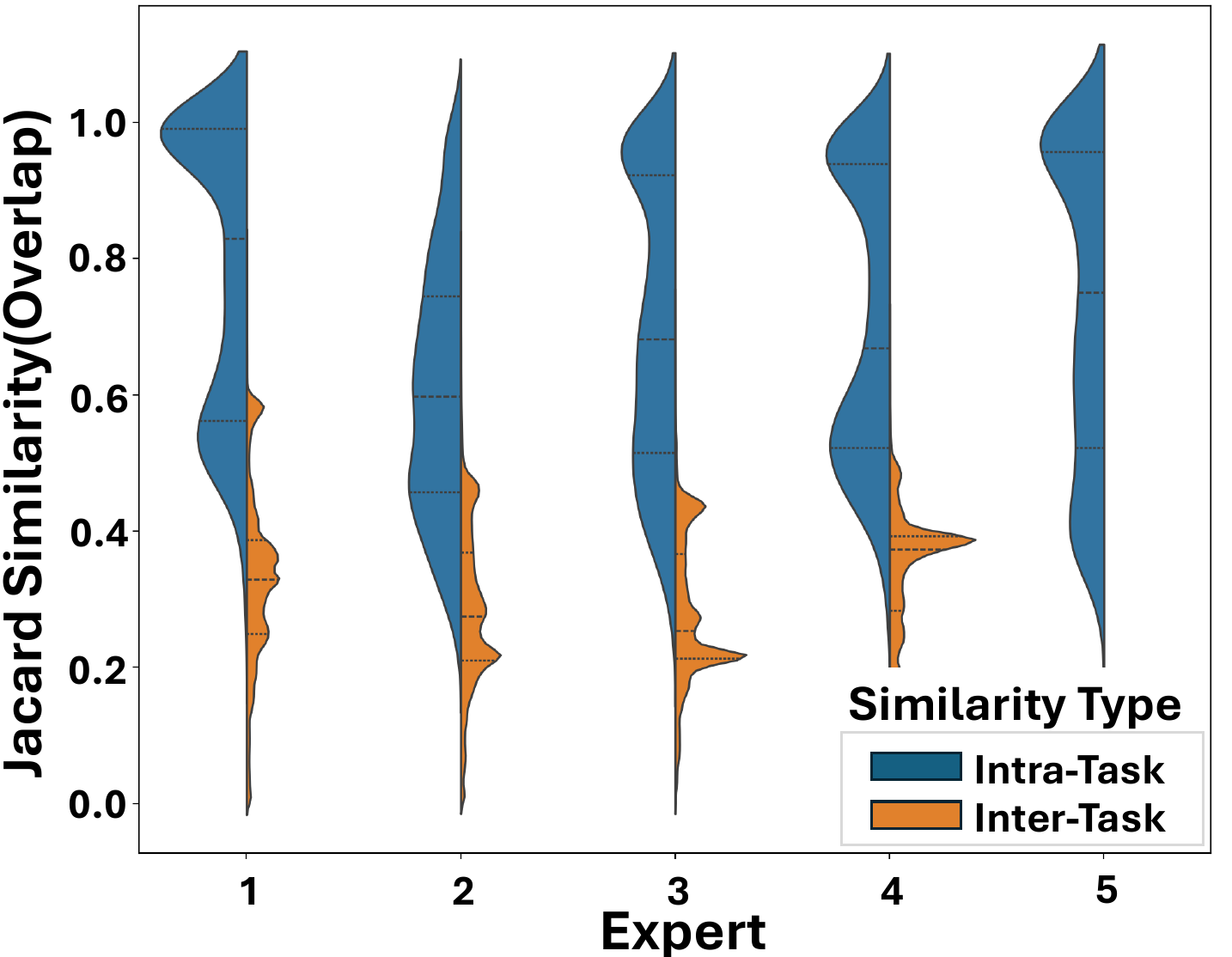}
    \end{minipage}
    \caption{
      \textbf{DG-Gated Routing is Enabled by Effective Feature Separation.} 
      The Class-IL routing matrix shows that inputs are correctly routed to their corresponding expert. For example, inputs from classes 0 and 1 (Task 0) are primarily routed to Expert 0. This routing is enabled by strong pattern separation in the DG layer, where the Jaccard similarity between patterns from different tasks ('Inter-Task') is significantly lower than within the same task ('Intra-Task').
    }
    \vspace{-3mm}
    \label{fig:gating_mechanism}
\end{figure}

\subsection{Training Details}
We use a 2‐conv + 2‐FC LeNet backbone for CIFAR‐10 and and a 4-conv + 2-FC CNN for Tiny-ImageNet (conv layers: 64→128→256→512 filters with 3×3 kernels and max pooling).  The GridCellLayer has $M=4$ sinusoidal units, DG size $=1024$ with top‐$k$ sparsity $k=5\%$, CA3/CA1 MLP widths $(512,256)$ and $(512,256,128)$ respectively, and replay buffer of 200 samples per task.  Models are trained with Adam for 10 epochs per CIFAR‐10 task (20 for Tiny‐ImageNet), $\lambda_{\mathrm{ewc}}=0.1$, and contrastive margin $m=0.2$.  

Table~\ref{tab:mega-results-flops-simple} reports Task‑IL accuracy, Class‑IL accuracy, and inference FLOPs per forward pass for both Small and Large HiCL variants alongside a suite of baselines. We fix the buffer at 200 or 500 samples per task and evaluate on Split CIFAR‑10 (columns 4–6) and Split Tiny‑ImageNet (columns 7–9). Baselines are grouped by methodology: regularization‑based, replay‑based, architectural/hybrid, and the joint offline upper bound.

\section{Discussion}
\label{sec:discussion}

\subsection{Summary of Key Results}
HiCL demonstrates that hippocampal-inspired architecture can achieve state-of-the-art continual-learning performance with high computational efficiency. A key advantage of our DG-gated design is its support for conditional computation, where the sparse DG codes select a single expert for activation per input, avoiding the need to execute all $N$ expert pipelines.

This approach drastically lowers the total computational footprint. For instance, on Split CIFAR-10, our Small model requires 10.81 MFLOPs for the shared backbone and 5.48 MFLOPs for a single expert, resulting in an efficient inference cost of only 16.3 MFLOPs. The Large model's cost under the same conditions is 49.7 MFLOPs. On the more complex Split Tiny-ImageNet, these costs are 49.1 MFLOPs for the Small model and 172.7 MFLOPs for the Large model. While activating a single expert may offer less GPU parallelism than a batch-parallel approach, the significant reduction in total operations makes HiCL well-suited for continual learning on resource-constrained hardware. Table~\ref{tab:mega-results-flops-simple} reports the full model capacity for fair comparison with baselines that execute all parameters, while conditional computation enables the 16.3 MFLOPs efficiency advantage described above.


\subsection{The Mechanism of DG-Gated Routing}

The success of our routing mechanism stems from effective pattern separation in the DG layer, as evidenced by our ablation study (Table~\ref{tab:ablation}). Removing DG gating or the contrastive loss collapses Class-IL accuracy from 57.99 to approximately 19, while omitting top-$k$ sparsity reduces Task-IL accuracy from 92.67 to 55.85, demonstrating that sparse separation is critical to the model's performance.

Each expert's Dentate Gyrus (DG) module learns a unique feature vocabulary tailored to its specific task. For an expert trained on Task 1 (e.g., distinguishing cats from dogs), its DG module becomes highly attuned to features like fur texture, whiskers, and ear shape. In contrast, an expert trained on Task 2 (e.g., airplanes vs. ships) will learn to amplify features related to wings, fuselages, and water.

The expert's prototype or abstraction, which is the average of all its DG codes, becomes a compact summary of this specialized vocabulary. When a new image of a cat is presented to the model, it is processed by all experts' DG modules in parallel. However, only the DG module from the ``cat/dog'' expert will produce a strong and characteristic sparse activation, as the input's features strongly resonate with its learned vocabulary. The other experts, lacking this specialization, will produce weaker and less distinct patterns.
\begin{figure}[!ht]
  \centering
  \textbf{CIFAR-10 t-SNE}\\
  \hspace{-10mm}\includegraphics[width=0.6\linewidth]{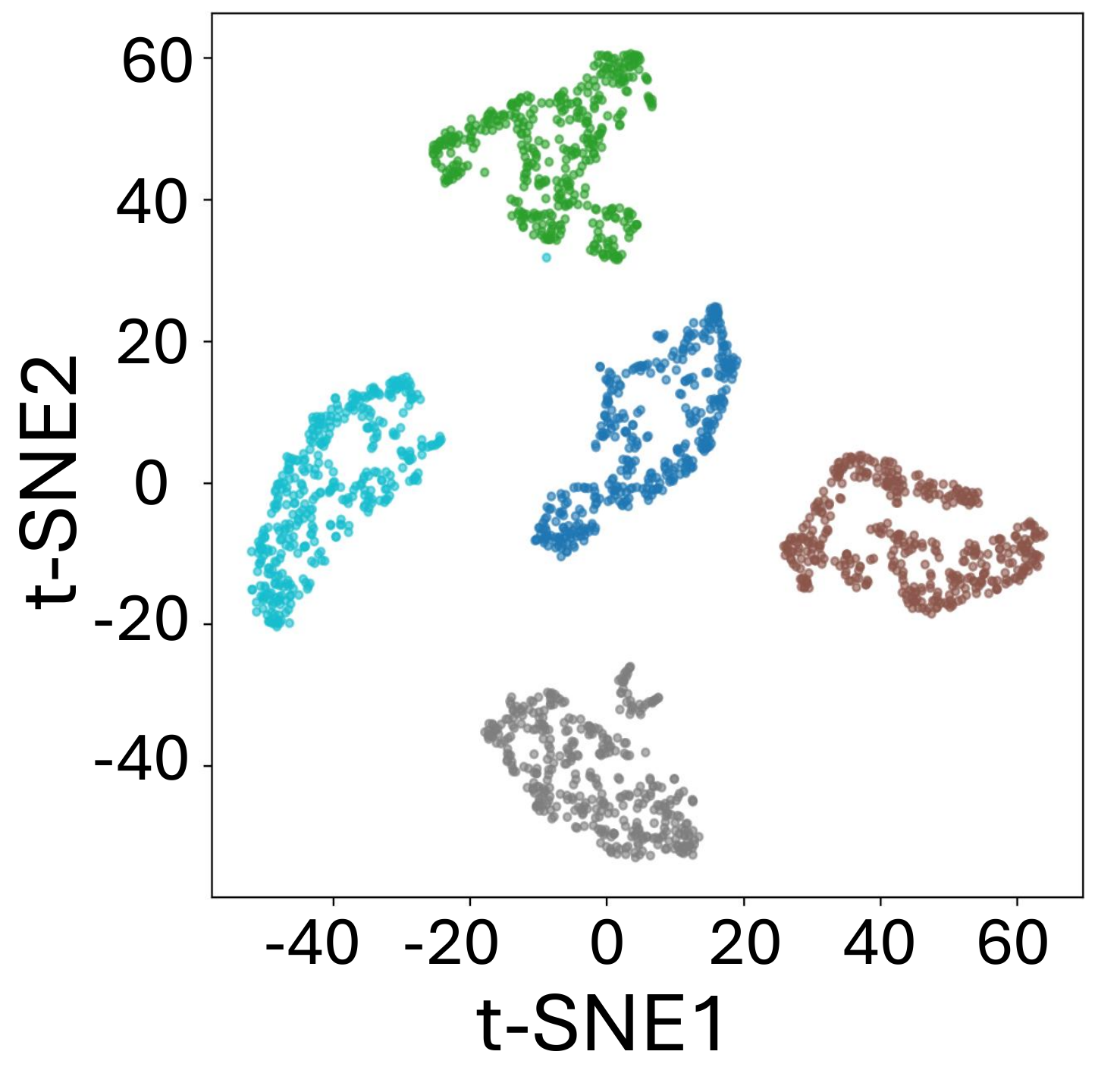}
  \caption{
    \textbf{DG Feature Separation and Prototype Orthogonality.} t-SNE plot for Split CIFAR-10.
  }
  \label{fig:dg_separation_cifar}
\end{figure}

\begin{figure}[!ht]
  \centering
  \textbf{Tiny-ImageNet t-SNE}\\
  \hspace{-10mm}\includegraphics[width=0.6\linewidth]{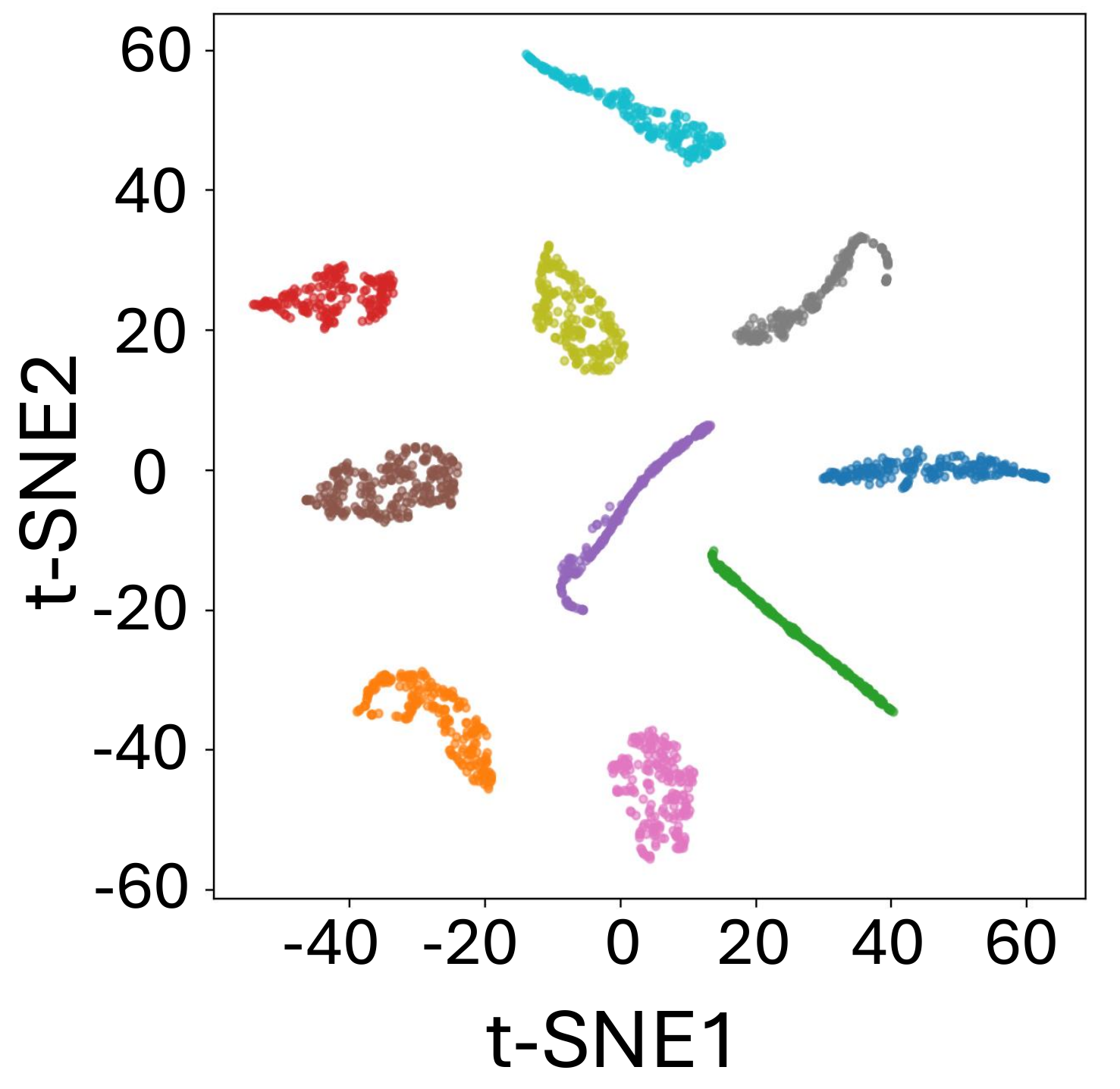}
  \caption{
    \textbf{DG Feature Separation and Prototype Orthogonality.} t-SNE plot for Split Tiny-ImageNet.
  }
  \label{fig:dg_separation_tiny}
\end{figure}

This separation is illustrated by the two dimensional t-SNE embeddings in Figures~\ref{fig:dg_separation_cifar} and \ref{fig:dg_separation_tiny}. Each colored group corresponds to a learned prototype, and the points form compact clusters with clear gaps between groups. The minimal overlap across clusters indicates that prototypes associated with different tasks occupy distinct regions of the representation space, which is consistent with low cross task similarity and reduced interference during continual learning.
Therefore, the routing decision, a simple cosine similarity check against the prototypes is not just a superficial pattern match. It is a measure of \textbf{representational resonance}. The input is naturally routed to the expert whose internal feature vocabulary is best suited to understand and process it, a choice made simple and reliable by the clear, quantifiable separation.

\subsection{Memory Robustness}
As shown in Table~\ref{tab:memory_accuracy}, HiCL degrades gracefully as the replay buffer size shrinks. For instance, reducing the buffer from 500 to 100 examples per task causes Task-IL accuracy to drop by only 2.7 percentage points (from 92.8\% to 90.1\%). This robustness stems from the DG’s pattern separation: each stored exemplar occupies a distinct sparse subspace, maximizing its utility during replay. Moreover, the model's feature-distillation and EWC regularization further stabilize representations, mitigating the effect of limited memory. These results suggest that efficient episodic encoding—rather than large buffer capacity—is a key driver of continual-learning performance.

\begin{table}[h]
\centering
\begin{tabular}{@{}lccc@{}}
\toprule
\textbf{Memory Size} & \textbf{Task-IL} & \textbf{Class-IL} & \textbf{Routing} \\
\midrule
100 & 90.1 & 53.4 & 56.0 \\
200 & 92.6 & 60.6 & 63.2 \\
300 & 91.5 & 63.1 & 65.6 \\
400 & 92.2 & 64.2 & 67.7 \\
500 & 92.8 & 65.4 & 67.8 \\
600 & 91.3 & 64.6 & 67.6 \\
1000 & 92.5 & 67.2 & 69.1 \\
5000 & 92.8 & 63.7 & 64.1 \\
\bottomrule
\end{tabular}
\caption{Accuracy vs Memory Size (Task-IL, Class-IL, Routing)}
\label{tab:memory_accuracy}
\end{table}

\begin{table}[h]
\centering
\begin{tabular}{@{}lcc@{}}
\toprule
\textbf{Configuration} & \textbf{Task-IL} & \textbf{Class-IL} \\
\midrule
Full HiCL & 92.67 & 57.99 \\
\midrule
\quad w/o Contrastive & 91.56 & 19.19  \\
\quad w/o DG Gating & 91.82 & 18.39 \\
\quad w/o top k Sparsity & 55.85 & 30.71  \\
\quad w/o CA1 & 92.64 & 52.39  \\
\quad w/o EMA & 90.63 & 56.29  \\
\quad w/o CA3 & 92.29 & 58.73  \\
\quad w/o EWC & 89.48 & 58.92  \\
\bottomrule
\end{tabular}
\small
\caption{Ablation study on Split CIFAR-10 (Replay szie = 200)}
\label{tab:ablation}
\end{table}

\subsection{Limitations and Failure Modes}
Despite its strengths, HiCL has several limitations. First, it operates under a Task IL assumption, requiring known task boundaries and labels during training. While Class IL inference is label free, Phase 1 still depends on task identities; fully unsupervised gating remains an open direction. Second, we use $\rho=0.05$ and $N=5/10$ experts across all tasks. These hyperparameters may not generalize well to domains with different levels of task heterogeneity or complexity. Third, routing can be sensitive in rare cases, with less than five percent of samples showing class confusion due to overlapping DG codes. This could potentially be addressed with adaptive gating thresholds or multi prototype routing. Lastly, our experiments are limited to vision datasets with LeNet style backbones. The model's performance on larger scale datasets or nonvisual domains such as language or reinforcement learning remains unexplored.

\subsection{Future Directions}
HiCL opens up several exciting paths for future research. A key direction is enabling unsupervised task discovery by leveraging clustering or contrastive pretraining on DG codes to infer task identities, moving toward fully label free Class IL. Another promising avenue is learning adaptive sparsity, where the sparsity ratio $\rho$ is dynamically adjusted per input or expert to balance representation separation and capacity as needed. For prototype maintenance, replacing exponential moving averages with more biologically grounded Hebbian updates or even transformer based attention could improve both interpretability and performance. Scaling HiCL to deeper architectures such as ResNet or Vision Transformers, and evaluating on larger benchmarks like ImageNet Split, will be essential to assess its effectiveness beyond small scale vision tasks. Finally, translating DG gating and expert routing into neuromorphic implementations using spiking neurons could enable real time, low power continual learning on edge hardware.



\newpage
\bibliography{aaai2026}

@article{cairney2018memory,
  title={Memory consolidation is linked to spindle-mediated information processing during sleep},
  author={Cairney, Scott A and El Marj, Nicole and Staresina, Bernhard P and others},
  journal={Current Biology},
  volume={28},
  number={6},
  pages={948--954},
  year={2018},
  publisher={Elsevier}
}

@article{demircigil2022sparse,
  title={A sparse quantized hopfield network for online-continual memory},
  author={Alonso, Nicholas and Krichmar, Jeffrey L},
  journal={Nature Communications},
  volume={15},
  number={1},
  pages={3722},
  year={2024},
  publisher={Nature Publishing Group UK London}
}

@article{kemker2018measuring,
  title={Measuring catastrophic forgetting in neural networks},
  author={Kemker, Ronald and McClure, Marc and Abitino, Angelina and Hayes, Tyler L and Kanan, Christopher},
  journal={Proceedings of the AAAI Conference on Artificial Intelligence},
  volume={32},
  number={1},
  year={2018}
}

@article{kirkpatrick2017overcoming,
  title={Overcoming catastrophic forgetting in neural networks},
  author={Kirkpatrick, James and Pascanu, Razvan and Rabinowitz, Neil and Veness, Joel and Desjardins, Guillaume and Rusu, Andrei A and Milan, Kieran and Quan, John and Ramalho, Tiago and Grabska-Barwinska, Agnieszka and others},
  journal={Proceedings of the National Academy of Sciences},
  volume={114},
  number={13},
  pages={3521--3526},
  year={2017},
  publisher={National Academy of Sciences}
}

@article{lee2020cortical,
  title={A cortical--hippocampal--cortical loop of information processing during memory consolidation},
  author={Rothschild, Gideon and Eban, Elad and Frank, Loren M},
  journal={Nature neuroscience},
  volume={20},
  number={2},
  pages={251--259},
  year={2017},
  publisher={Nature Publishing Group US New York}
}

@article{li2017learning,
  title={Learning without forgetting},
  author={Li, Zhizhong and Hoiem, Derek},
  journal={IEEE Transactions on Pattern Analysis and Machine Intelligence},
  volume={40},
  number={12},
  pages={2935--2947},
  year={2017},
  publisher={IEEE}
}

@article{riemer2018learning,
  title={Learning to learn without forgetting by maximizing transfer and minimizing interference},
  author={Riemer, Matthew and Cases, Ignacio and Ajemian, Robert and Liu, Miao and Rish, Irina and Tu, Yuhai and Tesauro, Gerald},
  journal={International Conference on Learning Representations},
  year={2018}
}

@article{rusu2016progressive,
  title={Progressive neural networks},
  author={Rusu, Andrei A and Rabinowitz, Neil C and Desjardins, Guillaume and Soyer, Hubert and Kirkpatrick, James and Kavukcuoglu, Koray and Pascanu, Razvan and Hadsell, Raia},
  journal={arXiv preprint arXiv:1606.04671},
  year={2016}
}

@article{schaul2016prioritized,
  title={Prioritized experience replay},
  author={Schaul, Tom and Quan, John and Antonoglou, Ioannis and Silver, David},
  journal={International Conference on Learning Representations},
  year={2016}
}

@article{schwarz2018progress,
  title={Progress \& compress: A scalable framework for continual learning},
  author={Schwarz, Jonathan and Czarnecki, Wojciech and Luketina, Jelena and Grabska-Barwinska, Agnieszka and Teh, Yee Whye and Pascanu, Razvan and Hadsell, Raia},
  journal={Proceedings of the 35th International Conference on Machine Learning},
  pages={4528--4537},
  year={2018}
}

@article{treves1994autoassociation,
  title={Computational Analysis of the Role of the Hippocampus in Memory},
  author={Treves, Alessandro and Rolls, Edmund T},
  journal={Hippocampus},
  volume={4},
  number={3},
  pages={374--391},
  year={1994},
  publisher={Wiley Online Library}
}

@article{wang2021triple,
  title={Triple-memory networks: A brain-inspired method for continual learning},
  author={Wang, Liyuan and Lei, Bo and Li, Qian and Su, Hang and Zhu, Jun and Zhong, Yi},
  journal={IEEE Transactions on Neural Networks and Learning Systems},
  volume={33},
  number={5},
  pages={1925--1934},
  year={2021},
  publisher={IEEE}
}

@article{hafting2005microstructure,
  title={Microstructure of a spatial map in the entorhinal cortex},
  author={Hafting, Torkel and Fyhn, Marianne and Molden, Sturla and Moser, May-Britt and Moser, Edvard I},
  journal={Nature},
  volume={436},
  number={7052},
  pages={801--806},
  year={2005},
  publisher={Nature Publishing Group}
}

@article{moser2008place,
  title={Place cells, grid cells, and the brain's spatial representation system},
  author={Moser, Edvard I and Kropff, Emilio and Moser, May-Britt},
  journal={Annual review of neuroscience},
  volume={31},
  pages={69--89},
  year={2008},
  publisher={Annual Reviews}
}

@article{rolls2013pattern,
  title={Pattern separation, pattern completion, and new inputs to the hippocampus},
  author={Rolls, Edmund T},
  journal={Neural Networks},
  volume={48},
  pages={5--14},
  year={2013},
  publisher={Elsevier}
}

@article{Rolls2013cortical,
title={The mechanisms for pattern completion and pattern separation in the hippocampus},
author={Edmund T. Rolls},
year={2013},
journal={Frontiers in Systems Neuroscience},
volume = {7},
pages = {74}
}

@article{petrantonakis2014compressed,
  title={A compressed sensing perspective of hippocampal function},
  author={Petrantonakis, Panagiotis C and Poirazi, Panayiota},
  journal={Frontiers in systems neuroscience},
  volume={8},
  pages={141},
  year={2014},
  publisher={Frontiers Media SA}
}

@misc{lisman2005recall,
  title={Memories of John Lisman},
  author={Otmakhova, Nonna A and Otmakhov, Nikolai and Griffith, Leslie C},
  year={2018},
  publisher={Frontiers Media SA}
}

@article{Mattar2018prioritized,
  title={Prioritized replay of salient experiences enables efficient learning in humans and machines},
  author={Mattar, Marcelo G and Daw, Nathaniel D},
  journal={Nature Human Behaviour},
  volume={2},
  number={10},
  pages={737--747},
  year={2018},
  publisher={Nature Publishing Group UK London},
  doi={10.1038/s41562-018-0437-8}
}

@article{gilbert2009role,
  title={The role of the CA3 hippocampal subregion in spatial memory: a process oriented behavioral assessment},
  author={Gilbert, Paul E and Brushfield, Andrea M},
  journal={Progress in Neuro-Psychopharmacology and Biological Psychiatry},
  volume={33},
  number={5},
  pages={774--781},
  year={2009},
  publisher={Elsevier}
}

@article{hainmueller2020dentate,
  author    = {Hainmueller, Thomas and Bartos, Marlene},
  title     = {Dentate gyrus circuits for encoding, retrieval and discrimination of episodic memories},
  journal   = {Nature Reviews Neuroscience},
  volume    = {21},
  number    = {3},
  pages     = {153--168},
  year      = {2020},
  month     = {Mar},
  doi       = {10.1038/s41583-019-0260-8},
  publisher = {Springer Nature}
}

@article{bartsch2011ca1,
  title={CA1 neurons in the human hippocampus are critical for autobiographical memory, mental time travel, and autonoetic consciousness},
  author={Bartsch, Thorsten and D{\"o}hring, Juliane and Rohr, Axel and Jansen, Olav and Deuschl, G{\"u}nther},
  journal={Proceedings of the National Academy of Sciences},
  volume={108},
  number={42},
  pages={17562--17567},
  year={2011},
  publisher={National Academy of Sciences}
}

@article{krizhevsky2009learning,
  title={Learning multiple layers of features from tiny images},
  author={Krizhevsky, Alex and Hinton, Geoffrey and others},
  year={2009},
  publisher={Toronto, ON, Canada}
}

@inproceedings{deng2009imagenet,
  title={Imagenet: A large-scale hierarchical image database},
  author={Deng, Jia and Dong, Wei and Socher, Richard and Li, Li-Jia and Li, Kai and Fei-Fei, Li},
  booktitle={2009 IEEE conference on computer vision and pattern recognition},
  pages={248--255},
  year={2009},
  organization={Ieee}
}

@article{yuksel2012twenty,
  title={Twenty years of mixture of experts},
  author={Yuksel, Seniha Esen and Wilson, Joseph N and Gader, Paul D},
  journal={IEEE transactions on neural networks and learning systems},
  volume={23},
  number={8},
  pages={1177--1193},
  year={2012},
  publisher={IEEE}
}

@article{cai2024survey,
  title={A survey on mixture of experts in large language models},
  author={Cai, Weilin and Jiang, Juyong and Wang, Fan and Tang, Jing and Kim, Sunghun and Huang, Jiayi},
  journal={IEEE Transactions on Knowledge and Data Engineering},
  year={2025},
  publisher={IEEE}
}

@article{guo2025deepseek,
  title={Deepseek-r1: Incentivizing reasoning capability in llms via reinforcement learning},
  author={Guo, Daya and Yang, Dejian and Zhang, Haowei and Song, Junxiao and Zhang, Ruoyu and Xu, Runxin and Zhu, Qihao and Ma, Shirong and Wang, Peiyi and Bi, Xiao and others},
  journal={arXiv preprint arXiv:2501.12948},
  year={2025}
}

@article{krotov2016dense,
  title={Dense associative memory for pattern recognition},
  author={Krotov, Dmitry and Hopfield, John J},
  journal={Advances in neural information processing systems},
  volume={29},
  year={2016}
}

@misc{hinton2022forward,
      title={The Forward-Forward Algorithm: Some Preliminary Investigations}, 
      author={Geoffrey Hinton},
      year={2022},
      eprint={2212.13345},
      archivePrefix={arXiv},
      primaryClass={cs.LG},
      url={https://arxiv.org/abs/2212.13345}, 
}

@inproceedings{skiers2024joint,
  title={Joint diffusion models in continual learning},
  author={Skier{\'s}, Pawe{\l} and Deja, Kamil},
  booktitle={Proceedings of the IEEE/CVF International Conference on Computer Vision},
  pages={4380--4390},
  year={2025}
}

@article{wang2024rehearsal,
  title={Rehearsal-free modular and compositional continual learning for language models},
  author={Wang, Mingyang and Adel, Heike and Lange, Lukas and Str{\"o}tgen, Jannik and Sch{\"u}tze, Hinrich},
  journal={arXiv preprint arXiv:2404.00790},
  year={2024}
}

@article{de2021continual,
  title={A continual learning survey: Defying forgetting in classification tasks},
  author={De Lange, Matthias and Aljundi, Rahaf and Masana, Marc and Parisot, Sarah and Jia, Xu and Leonardis, Ale{\v{s}} and Slabaugh, Gregory and Tuytelaars, Tinne},
  journal={IEEE transactions on pattern analysis and machine intelligence},
  volume={44},
  number={7},
  pages={3366--3385},
  year={2021},
  publisher={IEEE}
}

@inproceedings{madotto2020continual,
  title={Continual learning in task-oriented dialogue systems},
  author={Madotto, Andrea and Lin, Zhaojiang and Zhou, Zhenpeng and Moon, Seungwhan and Crook, Paul A and Liu, Bing and Yu, Zhou and Cho, Eunjoon and Fung, Pascale and Wang, Zhiguang},
  booktitle={Proceedings of the 2021 conference on empirical methods in natural language processing},
  pages={7452--7467},
  year={2021}
}

@article{zhang2022continual,
  title={Continual sequence generation with adaptive compositional modules},
  author={Zhang, Yanzhe and Wang, Xuezhi and Yang, Diyi},
  journal={arXiv preprint arXiv:2203.10652},
  year={2022}
}

@inproceedings{rebuffi2017icarl,
  title={icarl: Incremental classifier and representation learning},
  author={Rebuffi, Sylvestre-Alvise and Kolesnikov, Alexander and Sperl, Georg and Lampert, Christoph H},
  booktitle={Proceedings of the IEEE conference on Computer Vision and Pattern Recognition},
  pages={2001--2010},
  year={2017}
}

@article{rolnick2019experience,
  title={Experience replay for continual learning},
  author={Rolnick, David and Ahuja, Arun and Schwarz, Jonathan and Lillicrap, Timothy and Wayne, Gregory},
  journal={Advances in neural information processing systems},
  volume={32},
  year={2019}
}

@article{shin2017continual,
  title={Continual learning with deep generative replay},
  author={Shin, Hanul and Lee, Jung Kwon and Kim, Jaehong and Kim, Jiwon},
  journal={Advances in neural information processing systems},
  volume={30},
  year={2017}
}

@article{su2019generative,
  title={Generative memory for lifelong learning},
  author={Su, Xin and Guo, Shangqi and Tan, Tian and Chen, Feng},
  journal={IEEE transactions on neural networks and learning systems},
  volume={31},
  number={6},
  pages={1884--1898},
  year={2019},
  publisher={IEEE}
}

@article{mcclelland1995there,
  title={Why there are complementary learning systems in the hippocampus and neocortex: insights from the successes and failures of connectionist models of learning and memory.},
  author={McClelland, James L and McNaughton, Bruce L and O'Reilly, Randall C},
  journal={Psychological review},
  volume={102},
  number={3},
  pages={419},
  year={1995},
  publisher={American Psychological Association}
}

@article{Pham2021dualnet,
  title={Dualnet: Continual learning, fast and slow},
  author={Pham, Quang and Liu, Chenghao and Hoi, Steven},
  journal={Advances in Neural Information Processing Systems},
  volume={34},
  pages={16131--16144},
  year={2021}
}

@article{Kemker2018FearNet,
  title={Fearnet: Brain-inspired model for incremental learning},
  author={Kemker, Ronald and Kanan, Christopher},
  journal={arXiv preprint arXiv:1711.10563},
  year={2017}
}

@article{Banino2018Grid,
  title = {Vector-based navigation using grid-like representations in artificial agents},
  author = {Banino, Andrea and Barry, Caswell and Uria, Benigno and others},
  journal = {Nature},
  volume = {557},
  number = {7705},
  pages = {429--433},
  year = {2018}
}

@article{Leutgeb2007Pattern,
  title = {Pattern separation in the dentate gyrus and CA3 of the hippocampus},
  author = {Leutgeb, Stefan and Leutgeb, Jill K. and Treves, Alessandro and Moser, May-Britt and Moser, Edvard I.},
  journal = {Science},
  volume = {315},
  number = {5814},
  pages = {961--966},
  year = {2007}
}

@article{Hopfield1982Attractor,
  title = {Neural networks and physical systems with emergent collective computational abilities},
  author = {Hopfield, John J.},
  journal = {Proceedings of the National Academy of Sciences},
  volume = {79},
  number = {8},
  pages = {2554--2558},
  year = {1982}
}

@article{Shazeer2017MoE,
  title = {Outrageously Large Neural Networks: The Sparsely-Gated Mixture-of-Experts Layer},
  author = {Shazeer, Noam and Mirhoseini, Azalia and Maziarz, Krzysztof and others},
  journal = {arXiv preprint arXiv:1701.06538},
  year = {2017}
}

@inproceedings{Li2025MoETheory,
  title = {Theory on Mixture-of-Experts in Continual Learning},
  author = {Li, Hongbo and Lin, Sen and Duan, Lingjie and Liang, Yingbin and Shroff, Ness},
  booktitle = {International Conference on Learning Representations (ICLR)},
  year = {2025}
}

@article{Le2024PromptMoE,
  title={Mixture of experts meets prompt-based continual learning},
  author={Le, Minh and Nguyen, An and Nguyen, Huy and Nguyen, Trang and Pham, Trang and Van Ngo, Linh and Ho, Nhat},
  journal={Advances in Neural Information Processing Systems},
  volume={37},
  pages={119025--119062},
  year={2024}
}

@article{Caccia2022ContrastiveCL,
  author       = {Lucas Caccia and
                  Rahaf Aljundi and
                  Tinne Tuytelaars and
                  Joelle Pineau and
                  Eugene Belilovsky},
  title        = {Reducing Representation Drift in Online Continual Learning},
  journal      = {CoRR},
  volume       = {abs/2104.05025},
  year         = {2021},
  url          = {https://arxiv.org/abs/2104.05025},
  eprinttype    = {arXiv},
  eprint       = {2104.05025},
  bibsource    = {dblp computer science bibliography, https://dblp.org}
}

@article{Aghasanli2025Prototype,
  title = {Prototype-Based Continual Learning with Label-free Replay Buffer and Cluster Preservation Loss},
  author = {Aghasanli, Agil and Li, Yi and Angelov, Plamen},
  journal = {arXiv preprint arXiv:2504.07240},
  year = {2025}
}

@article{Nie2022EvoMoE,
  title = {EvoMoE: An Evolutional Mixture-of-Experts Training Framework via Dense-to-Sparse Gate},
  author = {Nie, Xiaonan and others},
  journal = {arXiv preprint arXiv:2112.14397},
  year = {2022}
}

@article{Fedus2022Switch,
  title = {Switch Transformers: Scaling to Trillion Parameter Models with Simple and Efficient Sparsity},
  author = {Fedus, William and Zoph, Barret and Shazeer, Noam},
  journal = {Journal of Machine Learning Research},
  volume = {23},
  number = {120},
  pages = {1--39},
  year = {2022}
}

@inproceedings{Zenke2017SI,
  title = {Continual Learning Through Synaptic Intelligence},
  author = {Zenke, Friedemann and Poole, Ben and Ganguli, Surya},
  booktitle = {ICML},
  pages = {3987--3995},
  year = {2017}
}

@inproceedings{Serra2018HAT,
  title = {Overcoming Catastrophic Forgetting with Hard Attention to the Task},
  author = {Serr{\`a}, Joan and Sur{\'\i}s, D{\'\i}dac and Miron, Marius and Karatzoglou, Alexandros},
  booktitle = {ICML},
  pages = {4548--4557},
  year = {2018}
}


\end{document}